\documentclass[letterpaper, 10 pt, conference]{ieeeconf}  

\IEEEoverridecommandlockouts                              

\usepackage{graphics} 
\usepackage{epsfig} 
\usepackage{amsmath} 
\usepackage{amsfonts}       
\usepackage{mathptmx} 
\usepackage{times} 
\newtheorem{definition}{Definition}[section]
\newtheorem{assum}{Assumption}[section]
\usepackage{multirow}
\usepackage{mathtools}
\usepackage{commath}
\usepackage{caption}
\usepackage{subcaption}
\usepackage{color}
\usepackage[ruled,vlined,linesnumbered]{algorithm2e}

\usepackage[utf8]{inputenc} 
\usepackage[T1]{fontenc}    
\usepackage{hyperref}       
\usepackage{url}            
\usepackage{booktabs}       
\usepackage{nicefrac}       
\usepackage{microtype}      

\SetKwInput{KwInput}{Input}
\SetKwInput{KwHyper}{Hyperparameters}
\SetKwInput{KwInitialize}{Initialize}
\SetKwInput{KwOutput}{Output}
\SetKwRepeat{Do}{do}{while}

\SetKwFunction{FunctionMain}{Main}
\SetKwProg{Fn}{Function}{:}{\KwRet $\pi_theta$}

\SetKwFunction{FunctionRL}{TrainOffPolicyRL}
\SetKwProg{RetPn}{Function}{:}{\KwRet $\theta$}

\SetKwFunction{FunctionTrainRL}{TrainRL}
\SetKwProg{Pn}{Function}{:}{}

\SetKwFunction{FunctionExpert}{ExpertGuidedRL}
\SetKwProg{Pn}{Function}{:}{}

\SetKwFunction{FunctionRLfinetune}{RLFineTune}
\SetKwProg{RetPn}{Function}{:}{\KwRet $\theta$}

\SetKwFunction{FunctionExpertActuation}{ExpertActuation}
\SetKwProg{RetPn}{Function}{:}{\KwRet $\theta$}

\SetKwFunction{FunctionRLControl}{SafeForRL}
\SetKwProg{RetPnThreshold}{Function}{:}{\KwRet $followRL$}

\SetKwFunction{FunctionGetPenalizedReward}{PenalizeReward}
\SetKwProg{RetPnPenalizedReward}{Function}{:}{\KwRet $r^p$}

\definecolor{dGreen}{rgb}{0.0, 0.5, 0.0}

\graphicspath{{figures/}}

\title{\LARGE \bf
A Joint Imitation-Reinforcement Learning Framework for Reduced Baseline Regret}

\author{Sheelabhadra Dey$^{1}$, Sumedh Pendurkar$^{1}$, Guni Sharon$^{1}$ and Josiah P. Hanna$^{2}$
\thanks{$^{1}$ Department of Computer Science and Engineering,
        Texas A\&M University, College Station, TX, USA
        {\tt\small \{sheelabhadra, sumedhpendurkar, guni\}@tamu.edu}}%
\thanks{$^{2}$ School of Informatics, University of Edinburgh, Edinburgh, United Kingdom
        {\tt\small josiah.hanna@ed.ac.uk}}
}

\begin{document}

\maketitle
\thispagestyle{empty}
\pagestyle{empty}

\begin{abstract}

%
In various control task domains, existing controllers provide a baseline level of performance that -- though possibly suboptimal -- should be maintained. 
%
%
Reinforcement learning (RL) algorithms that rely on extensive exploration of the state and action space can be used to optimize a control policy. However, fully exploratory RL algorithms may decrease performance below a baseline level during training.
%
In this paper, we address the issue of online optimization of a control policy while minimizing regret w.r.t a baseline policy performance.
We present a joint imitation-reinforcement learning framework, denoted JIRL. The learning process in JIRL assumes the availability of a baseline policy and is designed with two objectives in mind \textbf{(a)} leveraging the baseline's online demonstrations to minimize the regret w.r.t the baseline policy during training, and \textbf{(b)} eventually surpassing the baseline performance. JIRL addresses these objectives by initially learning to imitate the baseline policy and gradually shifting control from the baseline to an RL agent.
Experimental results show that JIRL effectively accomplishes the aforementioned objectives in several, continuous action-space domains. The results demonstrate that JIRL is comparable to a state-of-the-art algorithm in its final performance while incurring significantly lower baseline regret during training in all of the presented domains. Moreover, the results show a reduction factor of up to $21$ in baseline regret over a state-of-the-art baseline regret minimization approach.


\end{abstract}

\section{INTRODUCTION}

Deep reinforcement learning (RL) can produce policies that perform at, and even surpass, human-level control in various domains~\cite{mnih2015human}. As such, one might wonder why is deep RL not ubiquitously used to automate everyday tasks such as driving, traffic management, or medical procedures?
For such domains, it is necessary that the performance of any control policy is at least as good as the policy currently under operation and, ideally, improves upon it. 
%
Current RL algorithms, however, cannot provide such guarantees for the general case (unless assuming specific domain knowledge~\cite{ghavamzadeh2016safe}). As they possess no conceptual model of the world to begin with, such algorithms must perform extensive exploration, i.e., sampling different actions in various situations (world states). During exploration, the outcomes of different actions in different states are learned and the control function (denoted as `policy') is updated accordingly.
%
For example, consider an inefficient traffic signal controller at an intersection. Improving the controller's efficiency (e.g., w.r.t vehicle throughput) is desired however we should never allow the controller to perform significantly worse compared to the currently deployed controller as this might result in an abnormal cascading affect that can jam an entire city.



In RL, performance degradation over some baseline controller is commonly measured through the reduction in the accumulated reward and is denoted as \textit{baseline regret}~\cite{wu2017multi,ghavamzadeh2016safe}. Our main contribution is in proposing an approach for optimizing a control policy while minimizing baseline regret w.r.t a given baseline controller.

We propose the joint imitation-reinforcement learning (JIRL) framework for baseline regret minimization which utilizes a suboptimal policy that could be of any nature (deterministic/stochastic/rule-based/human-operated) while learning and applying an improved policy. Under this framework, a baseline policy and an RL policy jointly select actions. If the RL agent's action substantially differs from the baseline action then the baseline action is applied, otherwise the RL agent's action is applied. The RL policy is then updated with an off-policy learning algorithm while divergence from the baseline is penalized. Gradually, as the RL agent's policy improves, it is allowed to take actions that further diverge from the baseline. This procedure allows the RL controller to gradually find an optimal policy while discouraging highly-exploratory actuation. 
%
%
Note, however, that JIRL does not provide any guarantees regarding the resulting baseline regret. This is to be expected as providing such a guarantees is impossible without making specific assumptions about the environment.
Nonetheless, experimental results show a clear trend where the JIRL framework can fully train a real-world RC car in $45$ minutes leading to a $30\%$ improvement in performance w.r.t a baseline policy while minimizing baseline regret.



\section{PROBLEM DEFINITION}

We assume a Markov Decision Process~\cite{puterman2014markov} with state space, $S$, action space, $A$, transition probabilities, $P$, reward function, $R:S\times A \mapsto \mathbb{R}$, and discount factor, $\gamma$. An RL agent is assumed to start from  state $s_0$ and select action $a_0$ according to a policy, $\pi : S \mapsto A $. $\pi$ might be stochastic, i.e., mapping states to a distribution over actions. $\pi$ can be defined by a function approximator with a parameter set $\theta$ and is denoted $\pi_\theta$ in such cases.
Based on the chosen action, the agent receives a reward $r_0$ from the environment and reaches the next state, $s_1$, according to the transition probability, $P(s_1 | s_0, a_0)$. The process repeats and generates a trajectory, $\tau \coloneqq (s_0, a_0, r_0, s_1, a_1, r_1, s_2,\ldots)$. 
%

In addition to the standard MDP formulation, we assume an available baseline policy. 
%
%
%
%
%
\begin{assum}[Baseline Policy]
\label{assum:expert}
There exists a baseline policy, $\pi_b$, and at every time-step we can observe the action that $\pi_b$ would take in the current state.
\end{assum}
The baseline policy is not assumed to be optimal or exploratory (i.e., stochastic). That is, $\pi_b$ can be a deterministic (e.g., rule based) suboptimal controller. An example of such a controller is nowadays common traffic signal controllers at intersections.
%

\vspace{1.5mm}
\noindent \textbf{Objective:} \textit{Learn a parameterized policy that maximizes the sum of discounted rewards in an expected trajectory.}
\vspace{1.5mm}
That is, Maximize $J$ w.r.t. $\theta$ where:
\begin{equation}
    J(\pi_\theta) = \mathbb{E}_{\tau \sim \pi}\left[ \sum_t\gamma^t r_t \right]
    \label{eqn:objective1}
\end{equation}

\noindent \textbf{Desiderata:} \textit{Reduce baseline-regret during the training.} 

We generalize the definition of baseline regret that was used in previous work~\cite{wu2017multi,ghavamzadeh2016safe}.

\begin{definition}[Baseline Regret]
\label{def:bregret}
Given a baseline policy $\pi^b$, and a behavior policy $\pi$.
Define baseline regret as:
$$\mathcal{R(\pi)} = \mathbb{E}_{\tau \sim \pi, \tau^b \sim \pi^b} \left[ \sum_t  max( r^b_t-r_t, 0 ) \right]$$
where $r^b$ belongs to $\tau^b$ and $r$ to $\tau$
\end{definition}

A large body of work~\cite{sutton2018reinforcement} has focused on RL algorithms that are designed to efficiently optimize $J(\pi_\theta)$.
Furthermore, behavior cloning through supervised learning can be utilized to learn a policy from baseline demonstrations.
Such imitation learning approaches~\cite{ross2011reduction} can be applied offline, effectively eliminating baseline regret. However, when aiming to learn an optimized policy \textbf{and} minimize baseline regret, neither approach is sufficient as common RL algorithms perform extensive exploration and behavior cloning might learn from sub-optimal demonstrations, failing to optimize $J(\pi_\theta)$. Our proposed JIRL framework attempts to close this gap and merge the two approaches in a way that combines the best of each.

\subsection{Bounding the Baseline Regret}

Our JIRL framework does not provide any bounds on the accumulated baseline regret. This is, however, to be expected. In the general case i.e., with no additional assumptions regarding the provided MDP or baseline policy, the baseline regret cannot be bounded by a scalar. This claim follows from the fact that no regret bounds (including baseline regret) can be provided for the multi-armed bandit problem in the general case (see Bubeck et al.~\cite{bubeck2013bounded} for proof), and any multi-armed bandit instance (P1) can be reduced to an MDP (P2). The relevant reduction mapping constructs an MDP with a single state, $s_0$, in P2 where every bandit from P1, $b_i$, becomes an action at P2, $a_i$. The transition probabilities are $\forall a ~ P(s_0|s_0,a)=1$ and the reward, $R(s_0,a_i)$, for any action, $a_i$, in P2 follows the utility distribution from playing $b_i$ in P1. For such a construction, it is easy to see that, the baseline regret from following any policy in P1 over any baseline policy equals the baseline regret for equivalent policies in P2. 

\section{RELATED WORK}

A line of previous work did provide some guarantees regarding baseline regret. These, however, do not stand in contradiction to our previous claim regarding the infeasibility of regret bounds as these works all rely on some simplifying assumptions. Approaches assuming extensive baseline exploration relied on a stochastic baseline policy to sample high-variance trajectories. These trajectories were used to create a batch of data and then employed offline-RL or batch-RL~\cite{lange2012batch} algorithms for policy improvement~\cite{thomas2015high, laroche2019safe}. However, policy improvement is guaranteed only if the baseline policy executes an optimized trajectory with non-zero probability (baseline policy coverage) and the data generated is stationary. 
\cite{ghavamzadeh2016safe} additionally learned a model of the environment but assumed access to the error in the model estimation. JIRL, by contrast, can learn from a deterministic baseline policy and avoids learning a model due to the inaccuracies associated with model estimation without extensive exploration.

Adding safety constraints can prevent an agent from diverging from a baseline controller during and after training~\cite{geramifard2011uav, geramifard2012practical, achiam2017constrained}. However, ensuring such safety requires that for any achievable state, at least one safe action can be taken or the availability of a model that can accurately designate unavoidable safety violations following a given state-action pair. Algorithms that assume the availability of a safe action commonly rely on shielding~\cite{alshiekh2018safe}, action correction~\cite{zhang2016query, saunders2017trial, dalal2018safe, 8793742}, or ergodic MDPs~\cite{moldovan2012safe}. These approaches generally require domain knowledge about which actions will lead to constraint violations. Methods relying on a model assume that the model is available beforehand~\cite{berkenkamp2017safe, fulton2018safe} while others learn the model online~\cite{zhang2016query, dalal2018safe, wang2018safe, cheng2019end}. JIRL, by contrast, doesn't assume access to explicit safety constraints or the model of the environment.

%
In a different line of work, a number of approaches have aimed to ensure safe policy updates to improve upon a baseline policy on tabular problems~\cite{kakade2002approximately, pirotta2013safe}. By contrast, JIRL is evaluated on continuous action tasks where the value function and policy are parametrized via a function approximator (neural network). Trust region policy optimization (TRPO)~\cite{schulman2015trust} and proximal policy optimization (PPO)~\cite{schulman2017proximal} update stochastic policies by taking the largest step possible to improve performance, while satisfying a constraint on the KL-Divergence between the new and old policies. Since, TRPO and PPO approximate a monotonic improvement in performance, initializing TRPO (or PPO) with a baseline policy is a suitable candidate for comparison against JIRL in terms of the baseline regret.
In the TRPO/PPO framework, provided the domain allows stochastic policies, the initial stochastic baseline policy can be learned through observations (e.g., using imitation learning). Such an approach, however, requires a -- potentially expensive -- initial imitation learning phase. 
\section{THE JOINT IMITATION-REINFORCEMENT LEARNING FRAMEWORK}



In this section, we introduce our main contribution, the joint imitation-reinforcement learning framework (JIRL). 
JIRL extends over previous imitation and reinforcement learning algorithms by (1) generalizing the notion of penalized rewards~\cite{hester2018deep,DBLP:journals/corr/abs-1802-05313} to apply to continuous action spaces, (2) enabling learning from a baseline policy that can be queried during the training phase, and (3) defining a criterion for determining whether the RL agent or baseline policy should be given control at a particular time-step.

The JIRL framework is detailed in Algorithm~\ref{alg:JIRL}. At each timestep, the baseline policy is assumed to provide a suggested action to take, $a_t^b$, that is derived from its internal policy (Line~\ref{ln:ax}). 
Next, the RL agent determines if it should follow its own (stochastic) policy at the current state or defer to the baseline action (Line~\ref{ln:SafeRL}). 
If the RL policy is followed, $a_t^{rl}$ is sampled from $\pi_\theta$ and applied to the environment; otherwise, the baseline action, $a_t^{b}$ is applied.
Regardless of the action applied to the environment, the RL agent is trained on the observed outcome (given as a full transition $(s_t,a_t,r_t,s_{t+1})$, Line~\ref{ln:train}).
Note that training the RL agent on transitions originating from the baseline policy requires an off-policy RL learning procedure.
In cases where the baseline action $a_t^{b}$ was applied, we train the RL agent on a fabricated, counterfactual transition in which the RL agent took $a_t^{rl}$ and ended up in the same state that resulted from $a_t^{b}$ (Line~\ref{ln:train_fab}).
The reward for this fabricated transition -- obtained through the function \FunctionGetPenalizedReward{} -- is set to be lower than the reward affiliated with the baseline policy action. Doing so ensures that an RL agent (aiming to maximize return) will update its (stochastic) policy to shift probability towards $a_t^{b}$ from $a_t^{rl}$, i.e., towards imitating the baseline and reducing future baseline regret.

JIRL is, thus, a general framework that can work on top of any off-policy RL algorithm with a stochastic parameterized policy that implements \FunctionTrainRL{}. For example, an actor-critic algorithm~\cite{haarnoja2018soft} can implement \FunctionTrainRL{transition} as: store \textit{transition} in a replay buffer, periodically train both the actor and critic using stochastic gradient descent. JIRL, nonetheless, requires a specific implementation for \FunctionGetPenalizedReward{} in Line~\ref{ln:train_fab} and \FunctionRLControl{} in Line~\ref{ln:SafeRL}. These two functions determine the interplay between the imitation and reinforcement learning within JIRL and are discussed next.

\begin{algorithm}[h]
    \SetAlgoLined
\KwInput{baseline policy, $\pi_{b}$, maximum number of training steps, $\mathtt{L}$}
\KwOutput{optimized policy}
\KwInitialize{RL policy parameters, $\theta$}
    $s_0 \gets$ Reset environment\;
    \For{$t=0$ to $L$}{
            $a_{t}^{b} \leftarrow \pi_{b}(s_t)$; {\color{dGreen}\# Baseline's action}\\ \label{ln:ax}
            \uIf{\FunctionRLControl{$s_t, a_{t}^{b}, \pi_{\theta}, t$}}{ \label{ln:SafeRL}
                $a_{t} \sim \pi_{\theta}(\cdot|s_t)$; {\color{dGreen}\# RL action}\\ \label{ln:arl}
                }
            \Else{
                $a_{t} \leftarrow a_{t}^{b}$\;
                }
            $s_{t+1} \sim p(s_{t+1}|s_t, a_{t})$\;
            \FunctionTrainRL{$s_t, a_{t}, r_t, s_{t+1}$}\; \label{ln:train}
            \If{$a_{t} = a_{t}^{b}$}{
                {\color{dGreen}\# Fabricated (penalized) transition for $a^{rl}_{t}$}\\
                $a^{rl}_{t} \sim \pi_{\theta}(\cdot|s_t)$\;
                ${\color{red}r^p = \FunctionGetPenalizedReward{$r_t, a^b_t, a^{rl}_t$}}$\;
                \FunctionTrainRL{$s_t, a^{rl}_{t}, {\color{red}r^p}, s_{t+1}$}\; \label{ln:train_fab}
                }
            \If{$s_{t+1}$ is terminal}{
                $s_{t+1}=$ Reset environment\;
                }
        }
        \textbf{return} $\pi_\theta$
\caption{Joint Imitation-Reinforcement Learning}
\label{alg:JIRL}
\end{algorithm}

\subsection{Penalized reward for continuous actions}

Following~\cite{hester2018deep}, the reward for the fabricated transition (Line~\ref{ln:train_fab} in Algorithm~\ref{alg:JIRL}) is penalized such that the RL agent will be trained towards imitating the baseline policy.
We do so by introducing the fabricated transition, $(s_t,a_t^{rl},r^p_t,s_{t+1})$, where $r^p_t < r_t$. 
When considering continuous action spaces, setting a constant penalty, $l$, such that $r^p_t = r_t-l$~\cite{hester2018deep} will result in a non-smooth reward function as $\lim_{a \to a_t^{b}} R^p(s_t,a) \ne R^p(s_t,a_t^{b})$ where $R^p$ is the penalized reward function. As a result, an imitation learning process that uses fabricated transitions might fail to converge on the baseline policy. In order to address this issue, we present a continuous penalized reward function that is based on a Gaussian function. The penalized reward computation is presented in Algorithm~\ref{alg:penalty}. In this case, the penalized reward tends toward zero as $a^{rl}$ tends towards $a^{b}$. The hyperparameter representing the penalty function variance is chosen, for a given domain, based on the required action precision (smaller variance = more precise imitation). 

Note that for domains with a discrete action space, the penalty in  Algorithm~\ref{alg:penalty} should follow~\cite{piot2014boosted}, i.e., implemented as a margin function that is $0$ when $a_t^{rl} = a_t^{x}$ and positive otherwise.

\begin{algorithm}[]
    \SetAlgoLined
    \KwHyper{penalty variance, $\sigma^2$}
    \KwInput{reward, $r_t$, baseline policy's action, $a^{b}$, RL action, $a^{rl}$}
    \KwOutput{penalized reward value}
    \RetPnPenalizedReward{\FunctionGetPenalizedReward{$r$, $a^{b}$, $a^{rl}$}}{
    $\tilde{r} \leftarrow \abs{r}\left(1 - exp\left(-\norm{a^{b} - a^{rl}}/\sigma^2\right)\right)$\;
    $r^p \leftarrow r - \tilde{r}$\;
}
\caption{Penalized reward for fabricated transitions}
\label{alg:penalty}
\end{algorithm}



\subsection{RL control criteria}
\label{sec:safe}

\begin{figure*}[t]

\begin{center}
\begin{subfigure}{0.19\textwidth}
\includegraphics[width=\linewidth, height=2cm]{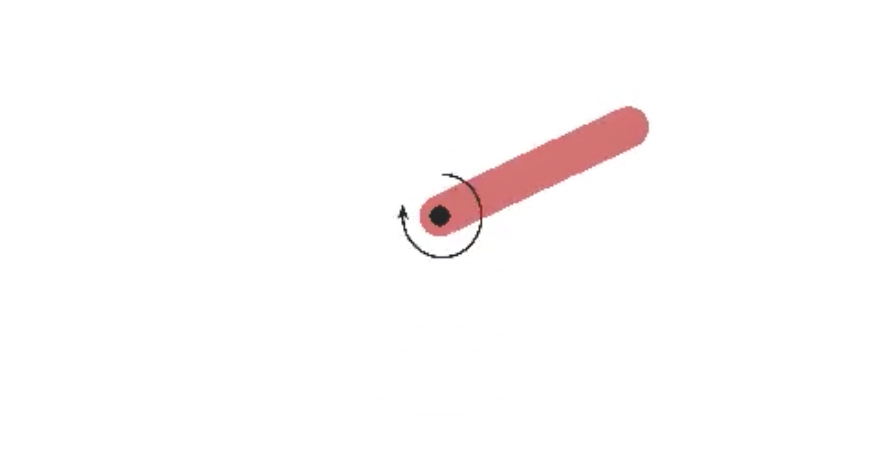} 
\caption{Inverted pendulum}
\end{subfigure}
\begin{subfigure}{0.19\textwidth}
\includegraphics[width=\linewidth, height=2cm]{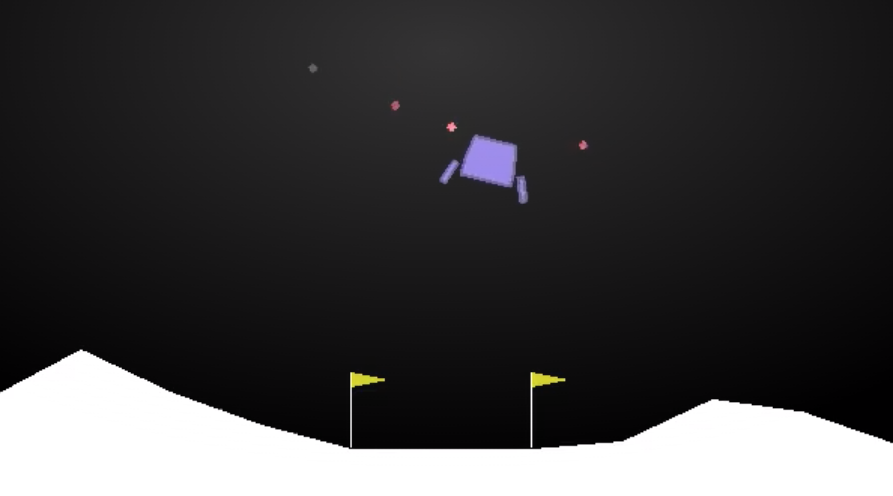}
\caption{Lunar lander}
\end{subfigure}
\begin{subfigure}{0.19\textwidth}
\includegraphics[width=\linewidth, height=2cm]{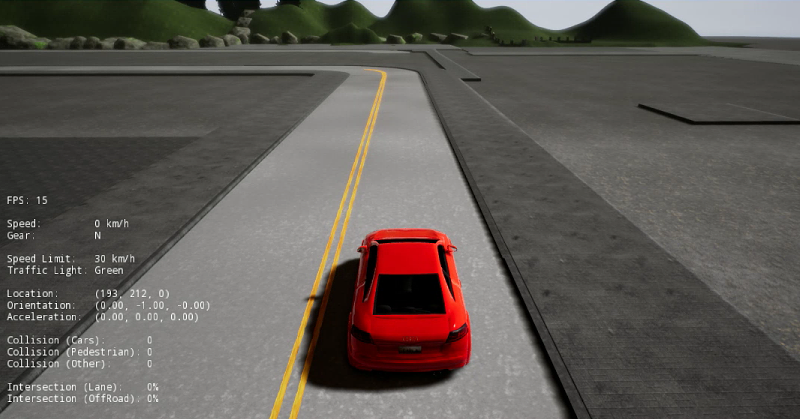}
\caption{CARLA}
\end{subfigure}
\begin{subfigure}{0.19\textwidth}
\includegraphics[width=\linewidth, height=2cm]{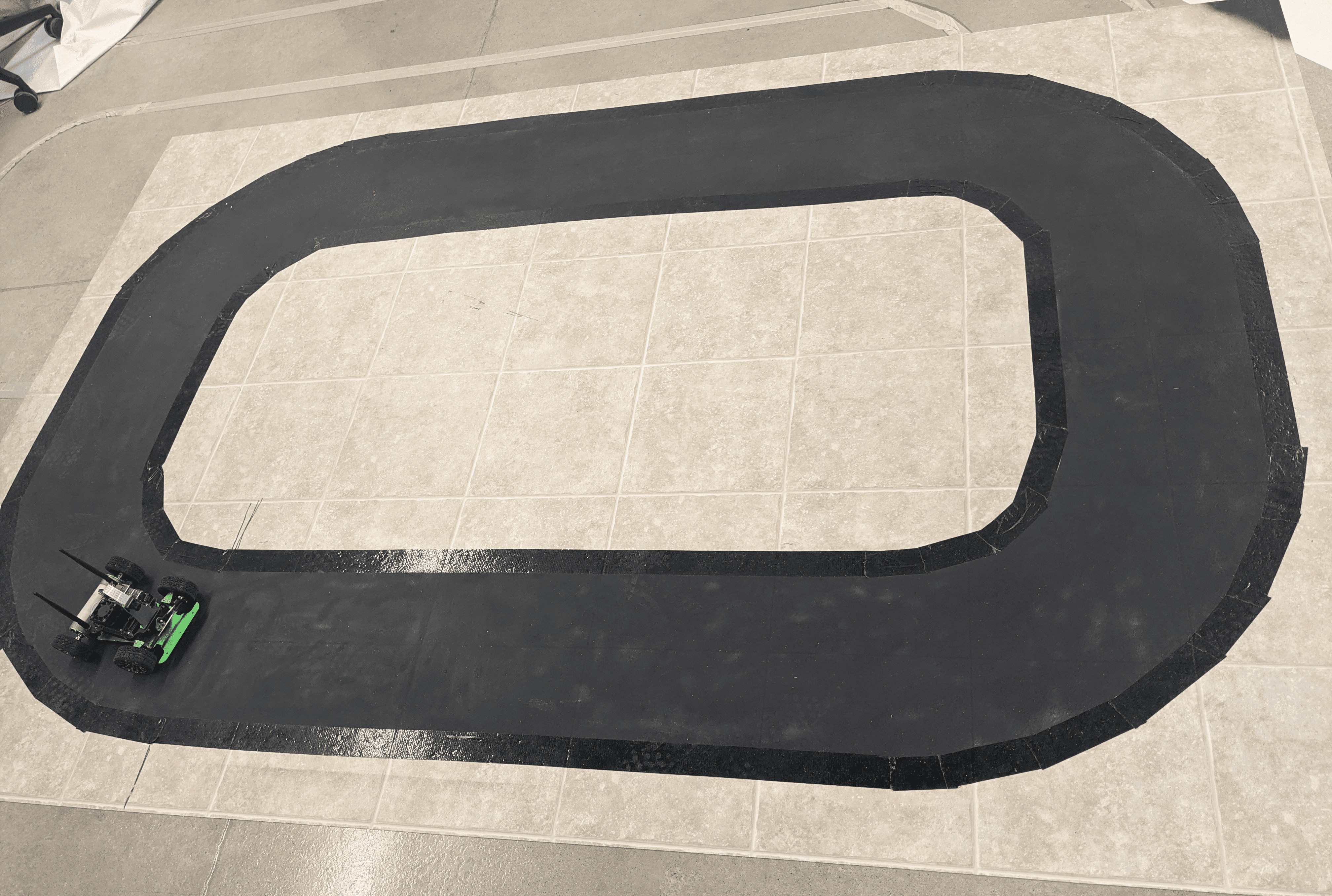}
\caption{JetRacer}
\end{subfigure}
\begin{subfigure}{0.19\textwidth}
\includegraphics[width=\linewidth, height=2cm]{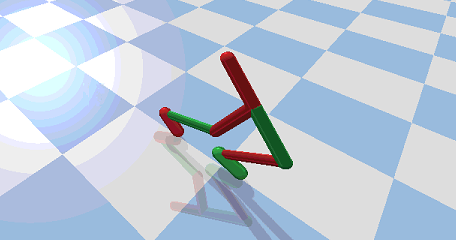}
\caption{Walker-2D}
\end{subfigure}
\caption{Snapshots of the domains}
\vspace{-5mm}
\label{fig:demo}
\end{center}
\end{figure*}


There is a balance to strike between imitating the baseline policy and taking exploratory actions so as to eventually outperform the same (presumably suboptimal) baseline. We address this balance through the function \FunctionRLControl{} through which the RL agent determines whether to act and explore or follow the baseline action and train towards imitating the baseline policy.
%
The intuition behind the selection criterion in \FunctionRLControl{} is that when the RL agent is uncertain about its actions (measured as high entropy policy), it should be trained towards imitating the baseline policy. As the RL agent becomes more certain regarding its actions, it is allowed to drift further away from the baseline policy and explore other promising actions. We define the \textit{divergence} value (denoted $\rho$) as the summation of the RL policy entropy term with the norm distance between the expected RL action and the baseline policy action.\footnote{For discrete action spaces, the norm distance between two actions, $a^{rl}, a^b$, can be defined as some constant if $a^{rl} \ne a^b$, else zero.}
%

Our proposed approach for determining which policy to follow is inspired by the control criteria that was presented in~\cite{menda2019ensembledagger}. Our suggested criteria is summarized in Algorithm~\ref{alg:safe}. It extends the criteria from~\cite{menda2019ensembledagger} by
considering divergence over consecutive time steps. Factoring the divergence values over several timesteps is important as small divergences can accumulate over time and result in a significant divergence. We observed this phenomenon when applying JIRL to an autonomous driving domain where the RL policy can slowly, yet steadily, steer the vehicle off the road. As a result, the divergence condition in Line~\ref{safeRL:ln:return} is factored over the minimum between the number of steps since the last RL-baseline control switch and a hyper-parameter $K$. An RL-baseline control switch occurs at time step $t$ if ($a_t=a^b_t$ and $a_{t-1}=a^{rl}_{t-1}$) or ($a_t=a^{rl}_t$ and $a_{t-1}=a^{b}_{t-1}$). $K$ is chosen appropriately for each domain. For domains where it is relatively easy to recover back to states the baseline would visit, e.g., Inverted Pendulum (see Section~\ref{sec:setup}), $K$ should be set lower. For domains that allow a chain of subtle divergences to lead to states that the baseline would not visit, e.g., lane following in autonomous driving, $K$ should be set higher. 

In some domains, it may be unreasonable to assume a baseline policy that provides an action at every time-step.
In such domains, JIRL can still be applied provided that the baseline policy can be queried any time that the baseline control is required. Autonomous vehicles with an expert safety driver are an example of such a domain. Such cases require minimal change to Algorithm~\ref{alg:JIRL} where if the baseline policy does not intervene (i.e., no action is provided) then $a^b_t \gets Null$ in Line~\ref{ln:ax}, and Algorithm~\ref{alg:safe} is simply implemented as $\textbf{return}~~ a^b_t$ equals $Null$.

\begin{algorithm}[]
    \SetAlgoLined
    \KwHyper{scaling factor, $\mathtt{C}$, number of steps to consider, $K$}
    \KwInput{state, $s$, baseline's action, $a^{b}$, RL policy, $\pi_\theta$, time step, $t$}
    \KwOutput{True iff the RL policy should be followed at the current state}
    \RetPnThreshold{\FunctionRLControl{$s$, $a^b$, $\pi_\theta$, $t$}}{
    $h \gets \mathcal{H}(\pi_\theta(\cdot|s)) $; {\color{dGreen}\# Entropy of the RL policy at the current state}\\
    $d \gets \norm{a^{b} - \mathbb{E}_{\pi_\theta}[A] }$; {\color{dGreen}\# Distance between the RL expected action and the baseline's action}\\ \label{ln:distance}
    $\rho_t \gets h + d$\;
    $k \gets $ minimum between $K$ and number of steps since last control switch\; 
    $followRL \gets \prod_{i=t-k}^{t}\pi_\theta(a_i^{b}|s_i) > \mathtt{C}\prod_{i=t-k}^{t}\rho_i$\; \label{safeRL:ln:return} 
}
\caption{Should we follow the RL policy?
}
\label{alg:safe}
\end{algorithm}




\subsection{Discussion}

It is important to note that, unless making specific assumptions regarding the implementation of \FunctionRLControl{}, the JIRL framework provides no bounds on the amount of baseline regret incurred.
%

Instead, JIRL provides a trade-off between baseline regret and allowable exploration towards finding an optimized final policy.
A lower value for the penalty variance (in Algorithm~\ref{alg:penalty}) encourages imitating the baseline -- reducing baseline regret but discouraging finding new, more optimal behaviors.
%
Similarly, a low scaling factor, $\mathtt{C}$, will allow the RL policy to take actions more frequently -- promoting exploration at the expense of possible higher baseline regret.


The underlying RL algorithm, implementing \FunctionTrainRL{}, is constantly being fed with a mixture of transitions generated from either the baseline, RL agent, or fabricated, penalized transitions.
There is an important distinction to make between these transitions. RL transitions can be utilized for on-policy learning towards the optimal policy. Baseline induced transitions can only be utilized for off-policy learning and the optimal policy can only be learned if \textit{coverage}~\cite[Chapter 5]{sutton2018reinforcement} is assumed. Fabricated transitions can be utilized for on-policy learning as they are composed of actions sampled from the RL policy. However, since the affiliated reward is fabricated, such transitions may bias learning away from the optimal policy. As a result, convergence to the optimal policy can be guaranteed if (1) the underlying RL algorithm provides such guarantees, and (2) only RL induced transitions are considered eventually.
As training progresses, the accumulated safe divergence value is expected to decrease since the RL policy entropy usually decreases (depending on the underlying RL algorithm). Consequently, RL control becomes more common and baseline/fabricated transitions are encountered less. As a result, learning the optimal policy requires an RL algorithm that ``forgets'' older transitions. Such a forgetting-attribute is common in RL algorithms that use a bounded replay buffer~\cite{haarnoja2018soft} which are, thus, particularly suitable for JIRL.

\section{EXPERIMENTS}

\begin{figure*}[t]
\begin{subfigure}{0.33\textwidth}
\includegraphics[width=\linewidth]{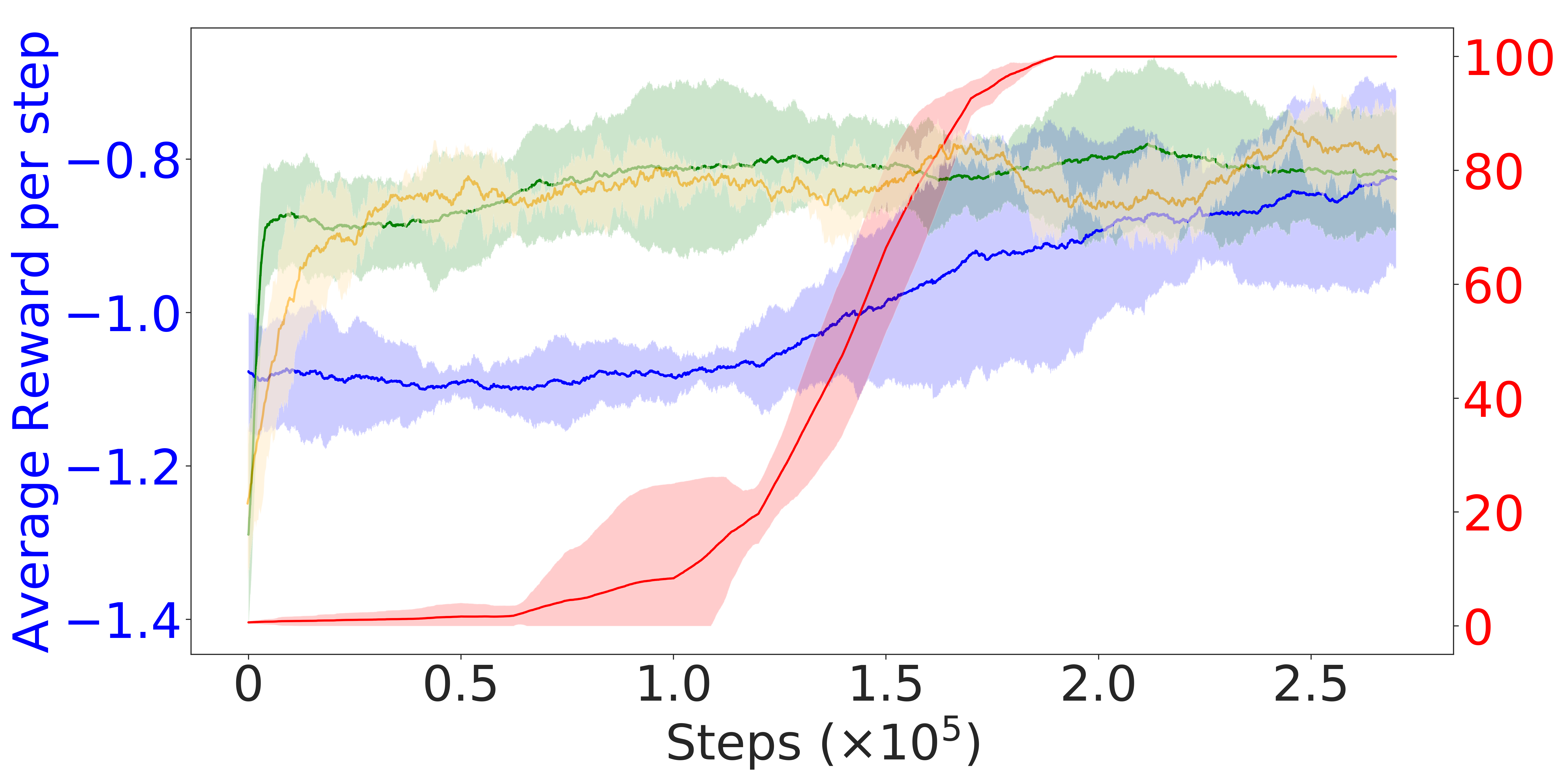} 
\caption{Inverted pendulum}
\label{fig:pendulum-reward}
\end{subfigure}
\begin{subfigure}{0.33\textwidth}
\includegraphics[width=\linewidth]{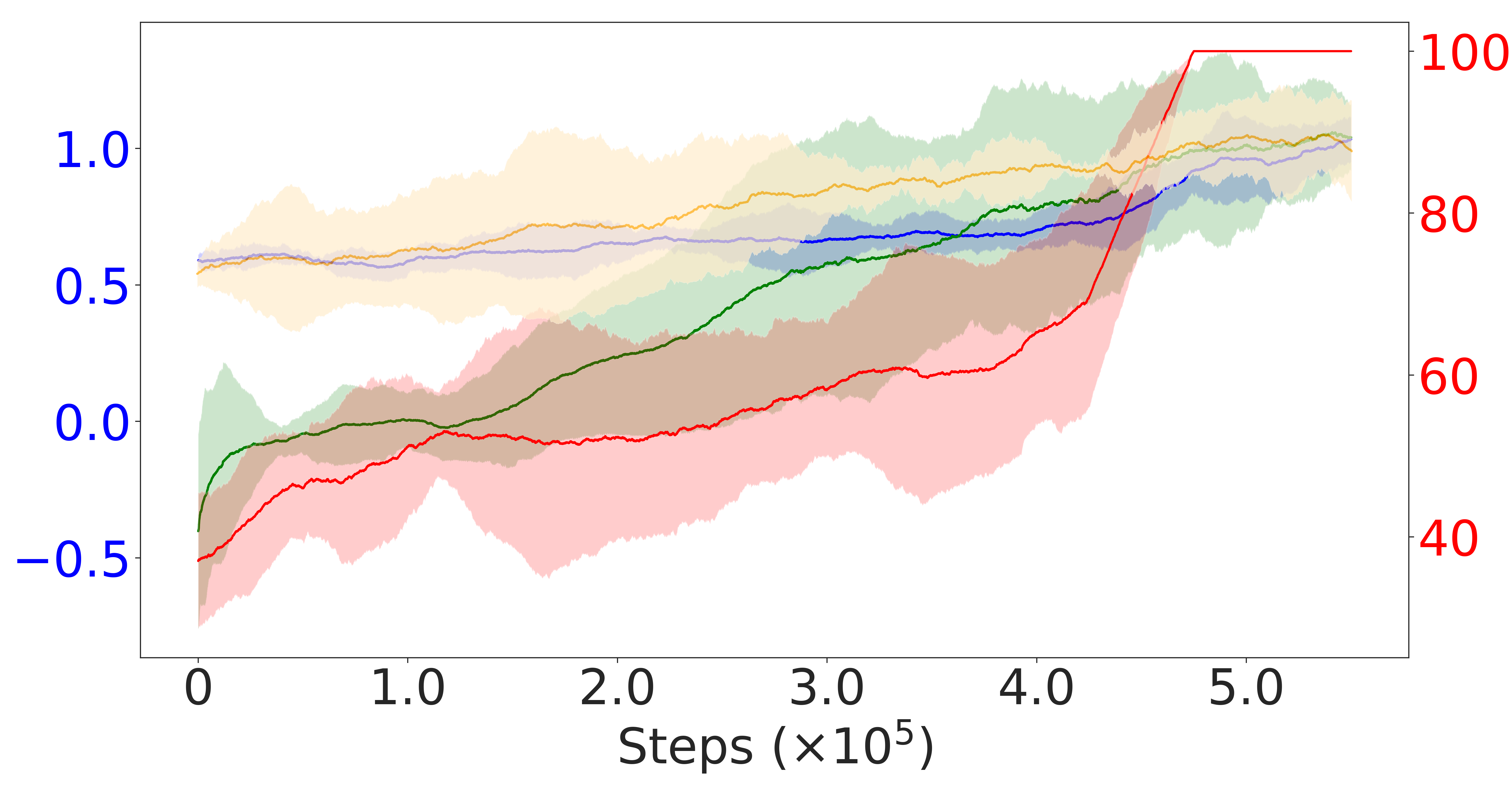}
\caption{Lunar lander}
\label{fig:lander-reward}
\end{subfigure}
\begin{subfigure}{0.33\textwidth}
\includegraphics[width=\linewidth]{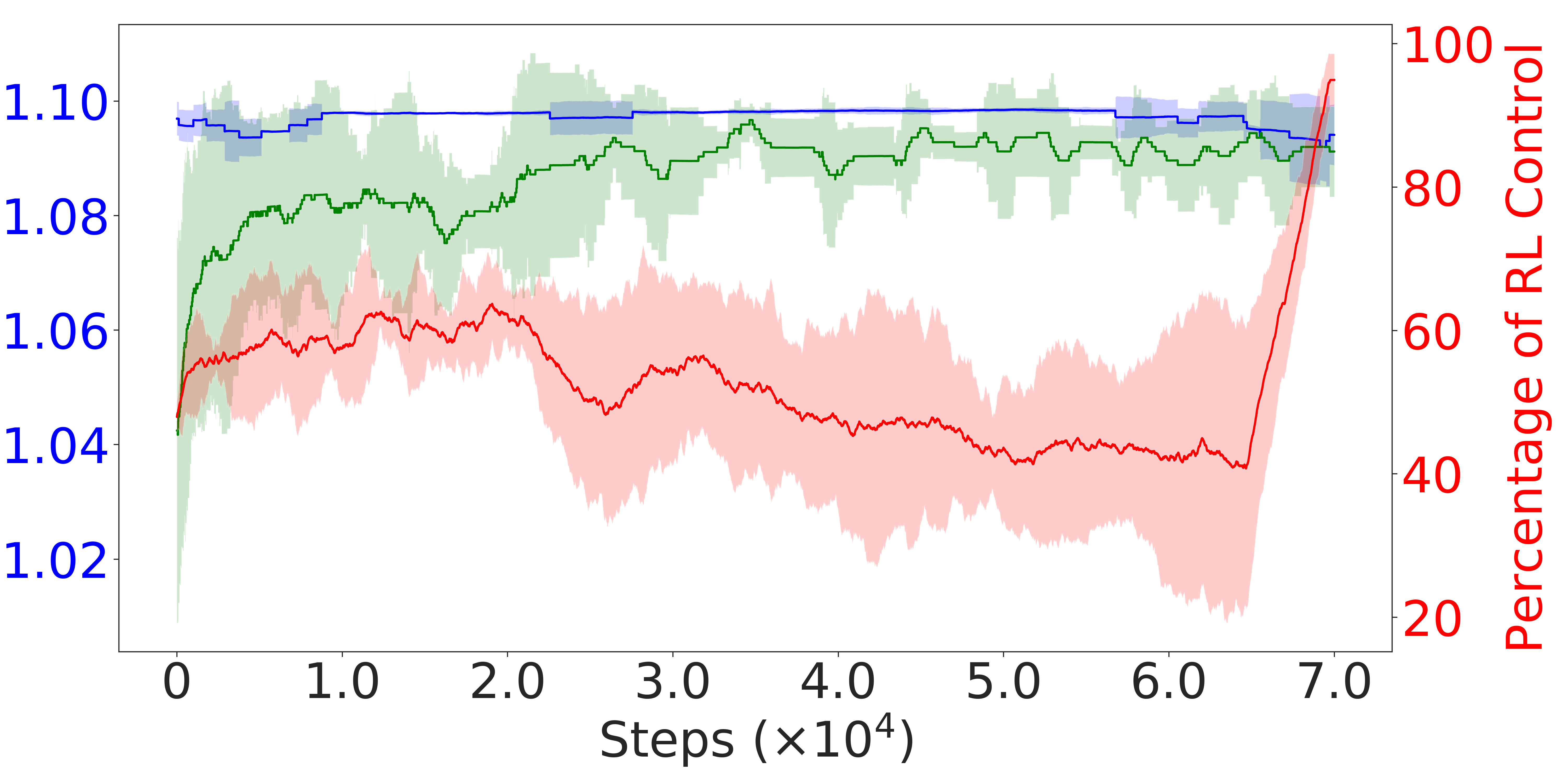}
\caption{Lane following (CARLA)}
\label{fig:carla-reward}
\end{subfigure}
\begin{subfigure}{0.33\textwidth}
\includegraphics[width=\linewidth]{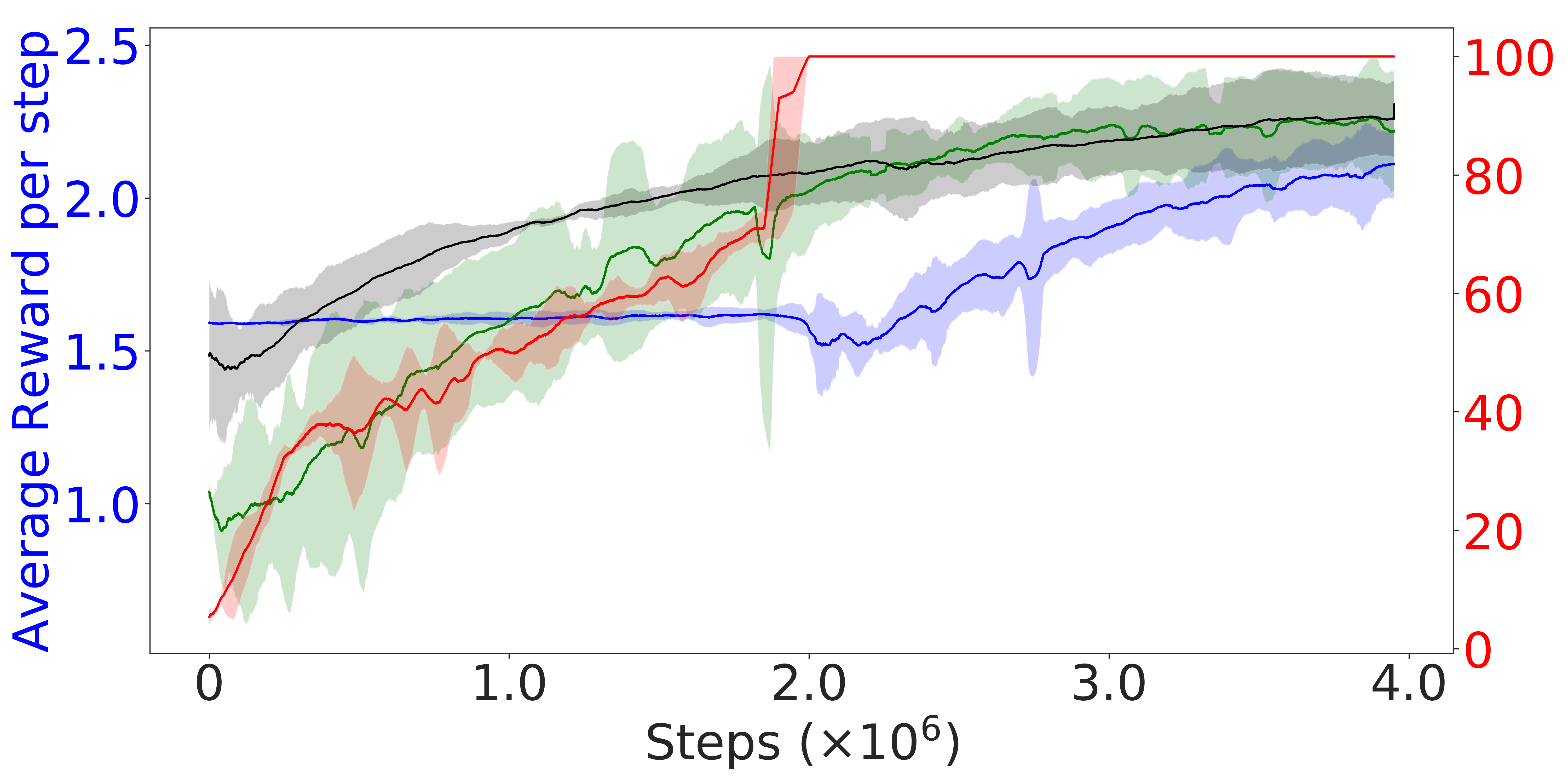} 
\caption{Walker-2D}
\label{fig:walker-reward}
\end{subfigure}
\begin{subfigure}{0.33\textwidth}
\includegraphics[width=\linewidth]{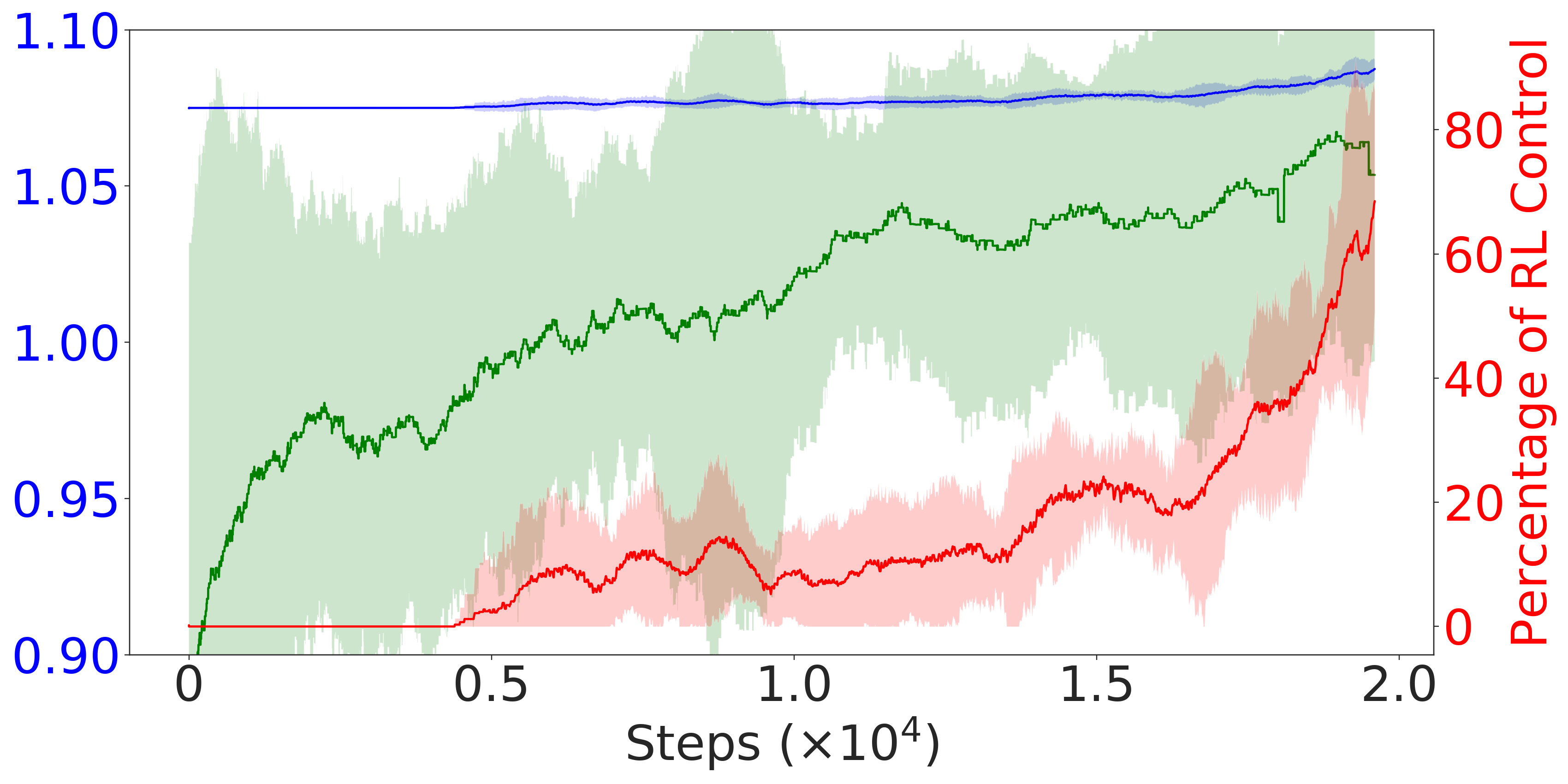} 
\caption{Lane following (JetRacer)}
\label{fig:rccar-reward}
\end{subfigure}
\begin{subfigure}{0.33\textwidth}
\includegraphics[width=\linewidth]{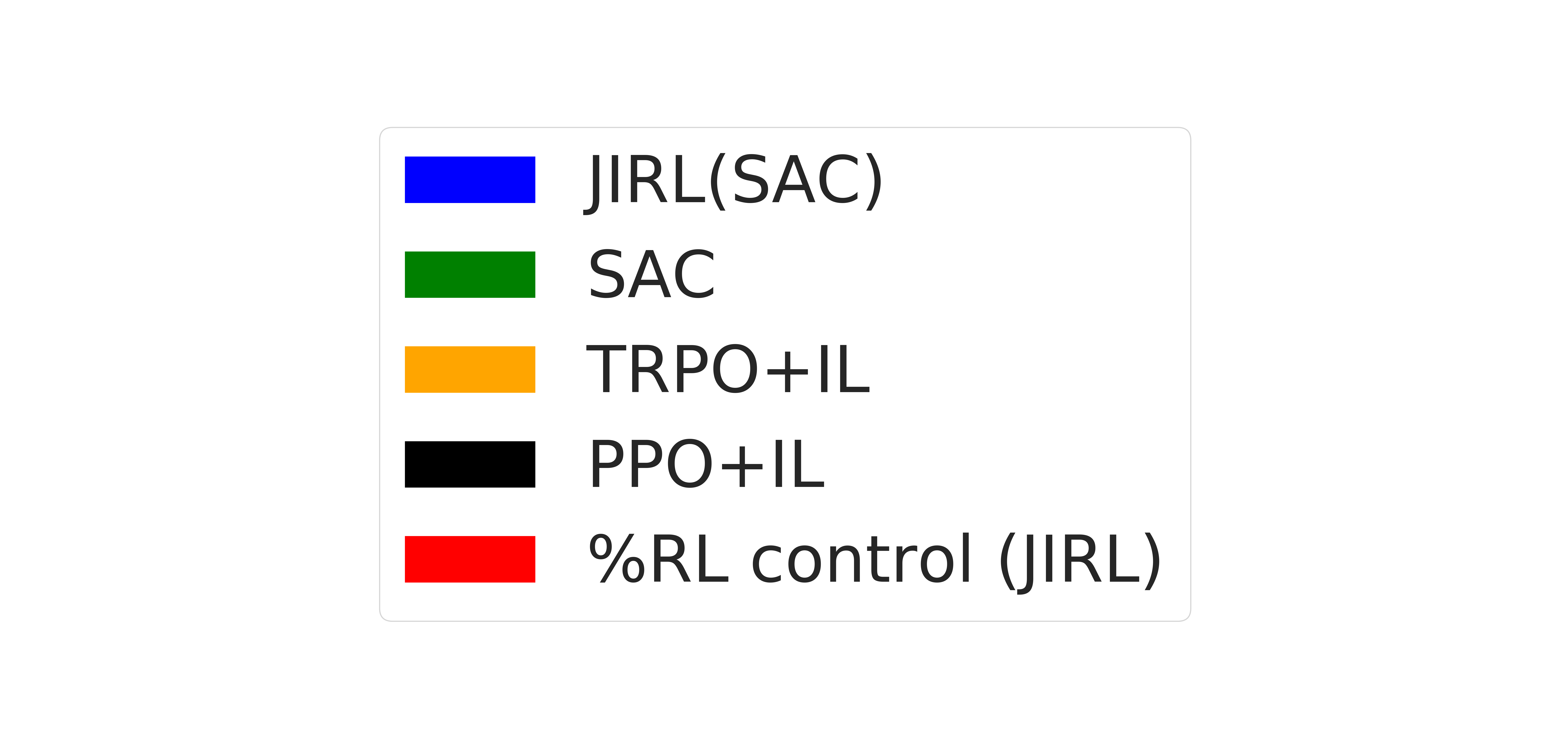} 
\caption{Plot legends}
\label{fig:lander-trpo}
\end{subfigure}

\caption{(a)-(e) Reward and percentage of RL control using the JIRL framework on top of SAC, vanilla SAC, and the best between TRPO+IL or PPO+IL when applicable. For the TRPO+IL and PPO+IL curves the required initial IL phase is omitted. In all the subfigures, the $x$-axis is the number of environment steps, the $y$-axis on the left is the smoothed reward, and the $y$-axis on the right is the percentage of RL control in the JIRL framework. The shaded region represents the 95\% confidence interval.} 
\label{fig:training-curves}
\end{figure*}



The goal of our experiments is to evaluate the effectiveness of JIRL in continuous control tasks with respect to the following objectives (a) leveraging the baseline's online demonstrations to reduce the regret w.r.t the baseline policy during training, and (b) eventually surpassing the baseline performance.
Specifically, we aim to show that (1) applying JIRL on top of a state-of-the-art RL algorithm results in significant reduction of the baseline regret while not degrading the final performance, and (2) JIRL outperforms a straightforward approach for eliminating baseline regret, which is, applying TRPO and PPO over the baseline policy (assuming a stochastic version is available or can be learned).

%

\subsection{Domain Description} \label{sec:setup}
Figure~\ref{fig:demo} shows snapshots from the domains used in our experiments. 
The goal in the \textbf{Inverted pendulum} task from the OpenAI gym~\cite{brockman2016openai} is to train an agent to swing up a pendulum and keep it at an upright position. In the \textbf{Lunar lander} task from the OpenAI gym, the objective is to train a space probe to land on a landing pad without crashing. The goal in the \textbf{Lane following (LF)} task is to train an autonomous vehicle to drive around a custom track following a lane. Using images from a front facing camera as input, we adopt the training set-up used in~\cite{drive-smoothly-in-minutes}. This domain is evaluated both in the CARLA simulator~\cite{Dosovitskiy17} and on a Waveshare JetRacer which is an autonomous scaled car that uses NVIDIA's Jetson Nano as the main control platform. In the \textbf{Walker-2D} task from the PyBullet environment~\cite{coumans2016pybullet}, a $2$-legged robot learns to stay upright and walk. Note that, for this domain, baseline regret can be significantly reduced when setting a higher $K$ value (as discussed in Section~\ref{sec:safe}). However, doing so also results in considerably slower learning.
We observed that values of $K$ between $5 - 10$ and $1 - 3$ achieve an acceptable trade-off between reducing the baseline regret and training time for the Lane following domain and the Walker-2D domain respectively. For the Inverted pendulum and Lunar lander domains, all values of $K$ between $1 - 5$ gave similar results in terms of reducing the baseline regret. $\sigma^2$ values from the range $[0.01 - 0.1]$ resulted in good performance (similar to those reported) in all domains. 

We observed that assigning a fixed penalty of $-1$ to transitions that resulted in the baseline given control (when the RL agent is deemed unsafe to act), yields slightly faster learning. Such events represent an unsafe divergence of the RL policy from the baseline and are, thus, penalized. such a penalty was used for obtaining the reported results.

For the Inverted pendulum, Lunar lander, Lane following (JetRacer) and Walker-2D tasks, sub-optimal deterministic baseline policies were defined in order to demonstrate that JIRL can learn from and outperform such a policy.
A detailed description of our experimental set-up and hyperparameters is available in a technical report that is available online at \url{https://pi-star-lab.github.io/JIRL}. The codebase for all the experiments is available at \url{https://github.com/Pi-Star-Lab/JIRL}. 

\subsection{Results}

In all the following experiments, soft actor-critic (SAC)~\cite{haarnoja2018soft} was used as the underlying RL algorithm within JIRL.
%
For each domain, we trained $5$ instances of JIRL(SAC) and vanilla SAC with the same set of hyper-parameters for SAC (as specified in the technical report) and different random seeds per instance. We used the implementation of SAC provided in Stable Baselines~\cite{stable-baselines}. Figure~\ref{fig:training-curves} shows the training curves for both approaches along with the TRPO/PPO baselines (when applicable).  

In all four domains, JIRL(SAC) resulted in a final policy that is at or above the vanilla SAC algorithm. Moreover, JIRL is shown to clearly reduce the baseline regret over vanilla SAC (the baseline performance can be seen on the left side of the JIRL curve where RL control is at a minimum). These results support the claim that applying JIRL on top of a state-of-the-art RL algorithm results in significant reduction of the baseline regret while not degrading the final performance.
In all but the Lane following (CARLA) domain, JIRL(SAC) resulted in a final policy that is superior to the baseline policy. The discrepancy in the Lane following (CARLA) domain is due to the use of a highly optimized baseline policy.
%
%
The Lane following (CARLA) baseline policy was trained using supervised learning on image-action pairs that were collected from an expert human demonstrator. SAC was unable to outperform these results when trained using a reward function similar to the one presented in~\cite{8793742}. When adjusting the reward function to reward greater speeds, it is possible to surpass the baseline performance (human demonstrator) by driving faster as demonstrated in the Lane following (JetRacer) domain (see Figure~\ref{fig:rccar-reward}). The JetRacer clocked a lap-time of $13.5$ seconds using the baseline policy while $45$ minutes of training using JIRL(SAC) reduced the average lap time to $9.4$ seconds (30\% improvement).

\vspace{1.5mm}
\noindent \textbf{Comparison with PPO and TRPO:}\\
\vspace{-3mm}

\begin{table}[t]
\centering
\begin{tabular}{l|ccccc}
    \toprule
    Domain  & \multicolumn{1}{p{0.8cm}}{\centering JIRL \\ 0-50\% RL}   &  \multicolumn{1}{p{0.75cm}}{\centering JIRL \\ 50-100\%}  & \multicolumn{1}{p{0.75cm}}{\centering JIRL \\ Full RL}  & \multicolumn{1}{p{0.75cm}}{\centering JIRL \\ Total} & 
    \multicolumn{1}{p{0.75cm}}{\centering TRPO/ \\ PPO}\\
    \midrule
    Inverted pendulum  & $3071$  & $396$  & $0$  & $3467$ & $37290$\\
    Lunar lander  & $2044$  & $819$  & $0$  & $2863$  & $60744$\\
    LF (CARLA)  & $13.6$  & $178.4$  & $0$  & $192$ & NA\\
    LF (JetRacer) &  $0.8$  &  $2.7$  &  $0$  &  $3.5$  & NA\\
    Walker-2D & $10$  & $188$  & $16616$  & $16814$  & $33039$\\
    \bottomrule
\end{tabular}
\caption{Comparison between the accumulated per step baseline regret in the JIRL framework and best of TRPO/PPO w.r.t the baseline performance. Results are averaged over $5$ runs. The advantage of JIRL over TRPO/PPO is statistically significant in all the domains based on a paired t-test.}
\vspace{-4.5mm}
\label{tab:fail}
\end{table}

We turn to compare JIRL(SAC) with an available baseline regret minimization approach, namely TRPO + imitation learning (IL) and PPO+IL over the baseline policy.
For each task, we considered both TRPO and PPO and include results of the algorithm that empirically performed better (PPO for Walker-2D and TRPO for Inverted Pendulum and Lunar Lander). In the Lane following tasks (CARLA and JetRacer), PPO+IL and TRPO+IL were not able to reach the same level of performance as the (highly optimized) baseline policy. Hence, PPO/TRPO results are omitted for this domain. 
The results in Figure~\ref{fig:training-curves} might seem favorable to PPO/TRPO, however the reader should remember that (1) PPO/TRPO requires an IL phase for learning a stochastic policy that is equivalent to the baseline policy. The long IL phase is omitted from these results as it significantly delays the RL phase (adding the IL phase would grow the x-axis for PPO/TRPO+IL by an order of magnitude); (2) PPO/TRPO, although presenting a fairly monotonic improvement in average performance, results in high performance variance which leads to high baseline regret; (3) due to low baseline policy coverage, PPO/TRPO might suffer from initial degradation in performance (see the Inverted Pendulum and Walker-2D domains); and (4) unlike JIRL(SAC), PPO/TRPO (using the specified reward function) were not able to reach the baseline performance in the Lane following domain.

Table~\ref{tab:fail} specifies the accumulated baseline regret in each task. The accumulated baseline regret for JIRL(SAC) is broken into the various training phases, where the phases are partitioned according to the percentage of RL control. The ``JIRL Full RL'' column corresponds to the phase starting when JIRL assigns 100\% RL control and ending when the average performance of JIRL is similar to that of SAC.
Empirically, we observed that JIRL(SAC) reduces the accumulated baseline regret over TRPO/PPO+IL by a factor ranging from $2$ in the Walker-2D domain to $21$ in the Lunar lander domain. JIRL(SAC) reduces the accumulated baseline regret over vanilla SAC by a factor ranging from $8$ in the Inverted pendulum domain to $400$ in the Lane following (JetRacer) domain (these results are not presented in Table~\ref{tab:fail} due to space constraints).

\section{CONCLUSION}

We introduce a joint imitation-reinforcement learning framework (JIRL) and demonstrate its ability to optimize a control policy while reducing regret with respect to an available baseline controller.
Assuming a baseline controller that is available during the training process, JIRL periodically switches between the baseline policy's actions and exploratory actions from the RL agent.
We define a control switching criterion that is based on the RL policy's entropy and its divergence from the baseline policy's actions accumulated over a series of timesteps.
Moreover, we presented a Gaussian penalty function for penalizing RL induced actions that lead to control switches.
Doing so, decreases the number of actions that lead to lesser reward than the baseline would accrue.
%
JIRL is shown to perform on par with the baseline during the learning process and eventually surpass a suboptimal baseline in all relevant domains.
Moreover, JIRL was shown to converge to a policy of similar quality (sum of discounted rewards) as a vanilla implementation of the underlying state-of-the-art RL algorithm (SAC) while reducing the accumulated baseline regret by up to $\times 21$.
%


\bibliography{references}

\end{document}